# Trustworthy AI: Deciding What to Decide

A Strategic Decision on Credit Default Swaps Investment

Caesar Wu[1], Yuan-Fang Li[2], Jian Li[3], Jingjing Xu[1] and Bouvry Pascal[1]

[1]University of Luxembourg, 6 Avenue de la Fonte L-4364 Esch-sur-Alzette, Luxembourg
[2] Monash University, 20 Exhibition Walk, Clayton, Vic 3800, Australia
[3] Dongbei University of Finance and Economics, Dalian, 116025, China
https://www.uni.lu/en/

**Abstract.** When engaging in strategic decision-making, we are frequently confronted with overwhelming information and data. The situation can be further complicated when certain pieces of evidence contradict each other or become paradoxical. The primary challenge is how to determine which information can be trusted when we adopt Artificial Intelligence (AI) systems for decision-making. This issue is known as "deciding what to decide" or Trustworthy AI. However, the AI system itself is often considered an opaque "black box". We propose a new approach to address this issue by introducing a novel framework of Trustworthy AI (TAI) encompassing three crucial components of AI: representation space, loss function, and optimizer. Each component is loosely coupled with four TAI properties. Altogether, the framework consists of twelve TAI properties. We aim to use this framework to conduct the TAI experiments by quantitive and qualitative research methods to satisfy TAI properties for the decision-making context. The framework allows us to formulate an optimal prediction model trained by the given dataset for applying the strategic investment decision of credit default swaps (CDS) in the technology sector. Finally, we provide our view of the future direction of TAI research.

## 1 Introduction

The notion of trust itself is a decision [1]. People often say, "Trust yourself", which is deciding what to decide. Similarly, the term trustworthy carries the same meaning. We often use these two terms interchangeably, but sometimes, they are quite confusing. Lexically, trust means belief in reliability. Trust is the result of something being perceived as trustworthy. These two terms form a complementary pair. Trustworthy Artificial Intelligence (AI) implies placing our belief in AI systems. The question is how to place our beliefs.

In his landmark book on the Peloponnesian War, Thucydides [2] argued that the vital difference between Sparta (winner) and Athens (loser) is the leadership quality that is

1.) Ability to process a massive amount of information, 2.) Quickly decide what to decide, and 3.) Carry action with a resolution. If we use AI/ML systems for strategic decision-making, two characteristics of leadership quality (1 & 2) are precisely the issue of Trustworthy AI. However, determining how much trust to place in a result generated by AI/ML can be daunting for many applications, such as financial investments, retail marketing, corporate planning, business strategies, public policy, and even health research. The challenge is how to frame TAI.

Perhaps Kissinger et al. [3] provided some clues for the solutions. They argued, "The AI, then, did not reach conclusions by reasoning as humans reason; it reached conclusions by applying the model it developed." In other words, the essence of AI/ML is to reverse the logic of human reasoning tradition. Instead of telling a machine what reasoning rules are, we tell the machine what we like. We can refer to it as a learning process. It consists of three essential components: 1.) representation space (models for values), 2.) loss function on data (data for evaluation), and 3.) optimizer (algorithms for selection). Determining how much trust for the AI/ML result is actually placing our trust in these components (See Fig. 1).

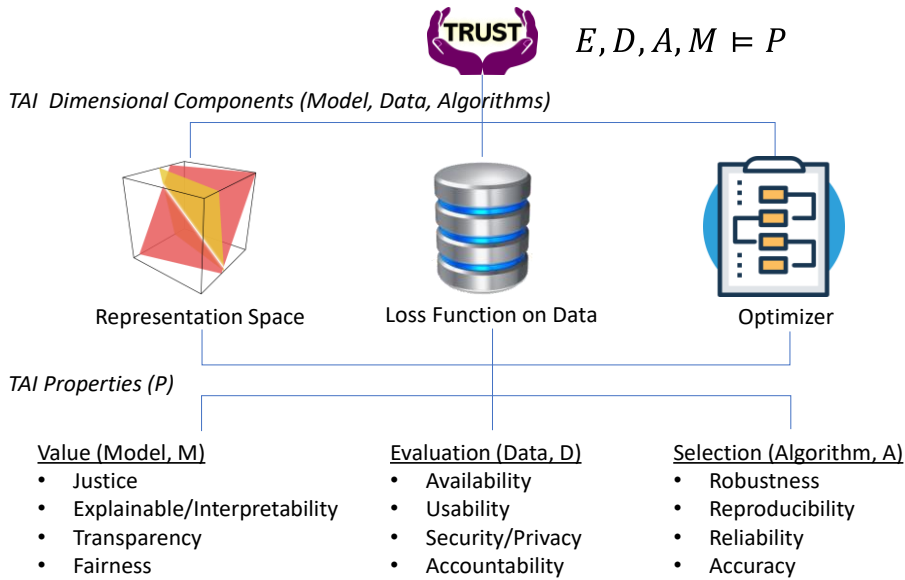

**Fig. 1.** Trustworthy AI Framework from a Strategic Decision-Making Perspective

If we move to the following components' level, twelve TAI properties underpin these components. Wing [4] proposed that models (M) and a system's environment (E) or data should be satisfied ($\vDash$) with a list of properties (P). Some researchers [5] have proposed actionable properties for TAI, such as moral operation, representation model, responsibility, and awareness of their morals. Others [7] classified TAI properties into three categories: technical, ethical and other requirements. We propose twelve TAI properties: justice, explainable/interpretability, transparency, fairness, availability, usability, security/privacy, accountability, robustness, reproducibility, reliability, and accuracy. They are organized into three groups (value, evaluation, and selection) loosely coupled with three learning components: representation space, loss function on data, and optimizer (See Fig. 1).

This framework demonstrates the relationship among TAI properties (P), learning components (M, D, and A), and TAI context or environments (E). For example, the representation space mainly aligns with the class of ethical value properties: justice, explainable/interpretability, transparency, and fairness. It is a box for machine learning programs to search for rules within the box. We want the learning result to satisfy our value systems. Therefore, the representation space has to meet the value properties, often ethical values or beliefs, even faith. When we design a loss function on data, a dataset must satisfy the TAI properties of accuracy, usability, security/privacy, accountability and data governance. [6] Likewise, the optimizer component should satisfy the properties of robustness, reproducibility, reliability and accountability. Overall, the decision context (E), data (D), selection algorithms (A), and representation space (M) should satisfy the TAI properties represented in Equation 1.

$$E, D, A, M \vDash P \qquad (1)$$

Generally, we may include some TAI properties and exclude others in a particular decision-making context. The decision context (E) and data (D) decide which TAI property (P) should be included and which one should be excluded in the detailed AI/ML process.

### 1.1 Research Question

Suppose we want to make a strategic investment decision regarding credit default swaps (CDS) for the technology sector in the financial derivative market (TAI context or environment E). The research question is, "What kind of model (M), dataset (D), and algorithms (A) will satisfy the listed TAI properties (P)?" Simply put, "How can we rely on the AI/ML result for an investment decision?

Why does this matter? If we remember the 2008 financial crisis, we know that the CDS were one of the primary sources for the 2008 crisis [12]. To a certain extent, the consequences of the 2008 crisis still impact global economics today. Moreover, we intend to generalize the TAI framework for a broad context of strategic decision-making applications.

### 1.2 Research Method

In order to solve this problem, we adopted the quantitative and qualitative research methods for this study. The quantitative methods [i.e. variable importance (VI), partial dependent plot (PDP)] focus on the dataset and the predictive model to satisfy explainability for the AI/ML results. The qualitative methods [i.e. individual conditional expectation (ICE) plot, local interpretable model-agnostic (LIME), and Shapley values (SHAP)] estimation for some important features is to interpret the details of the model for its transparency. By leveraging the research methods, we made the following contributions.

### 1.3 Main Contributions

We articulate a novel framework to handle many TAI issues, especially through explanation, interpretation, transparency, robustness, reproducibility, and accuracy. It is built upon the machine-learning components: 1.) representation space, 2.) loss function on data, and 3.) optimizer. We identified twelve TAI properties and grouped them into three categories. Each category is loosely coupled with each ML component. The framework allows us to address various TAI issues systematically.

We use GBM, Xgbm, and transformer models for the context of CDS prediction to demonstrate the new way of approaching the TAI issue. This study mainly focuses on the TAI's explanation, interpretation, and transparency properties via five techniques: VI, PDP, ICE plots, LIME, and SHAP.

We adopted Xgboost and transformer to build a predictive model for this decision context. Our experimental result indicated that Xgbm is much more compelling in satisfying some essential properties of transparency, interpretability, explainability, and reproducibility.

### 1.4 Scope of the Research

The rest of the paper is organized as follows: Section 2 is a literature survey that starts from types of representation space and loss function on data to optimizers regarding TAI properties. Section 3 introduces a bird's eye view of the dataset and experimental models. Section 4 is the experimental setup and results. Section 5 is the result analysis and discussion. Section 6 is the conclusion and future research direction.

## 2 Literature Review

The following literature review laid out a brief survey of five key related sub-topics regarding trustworthy AI: types of representation space, loss function on data, optimizer, CDS and strategic decision-making, and TAI techniques, especially on explainable AI (XAI). Based on our previous research experiences [36], we primarily focus on the gradient-boosting machine (GBM) or extreme gradient-boost machine (Xgbm) and transform techniques.

### 2.1 Types of Representation Space

The decision objectives determine how we build representation space (model) that allows a machine to search for rules effectively. Page [8] articulated seven types of models: reason, explain, design, communicate, act, predict and explore (or REDCAPE).

Kuhn and Silge [9] suggested only three types of models: descriptive, inferential, and predictive models. If we dive into details, these two taxonomies of models are similar. We can synchronize Page's classification and Kuhn's taxonomy, which descriptive models include "explain" and "communicate"; the inference model is associated with "reason" and "explore" while the predictive model means "design", "act", and "predict." (Refer to Fig. 2)

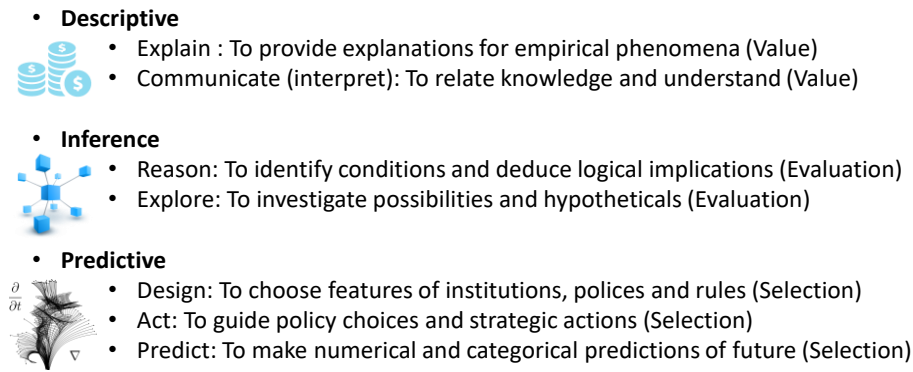

- **Descriptive**
  - Explain : To provide explanations for empirical phenomena (Value)
  - Communicate (interpret): To relate knowledge and understand (Value)

- **Inference**
  - Reason: To identify conditions and deduce logical implications (Evaluation)
  - Explore: To investigate possibilities and hypotheticals (Evaluation)

- **Predictive**
  - Design: To choose features of institutions, polices and rules (Selection)
  - Act: To guide policy choices and strategic actions (Selection)
  - Predict: To make numerical and categorical predictions of future (Selection)

**Fig. 2.** Types of Representation Space

**Descriptive models** aim to illustrate the characteristics of some data. They usually offer a trend or some clusters in the data. A typical example is customer segmentation [10]. If the dataset contains customer information about age, income, purchase history, gender, and ethnic group, the created model should reflect ethical values. **Inference models** often produce research questions or null hypotheses for further investigation. Exploratory data analysis (EDA) and feature selection are typical examples of an inference model for machine learning. The requirements of inference models are usable and available data that people can trust. The inference model aims to create better predictors. **Predictor models** often ask "what" rather than "how." They also provide a degree of uncertainty. Prediction is often closely related to explanation. The model choice depends on the problem context, the given data, and the performance requirements. The question of selecting a model leads to defining a loss function on data.

## 2.2 Loss Function on Data

The essence of the loss function on data is to quantify the discrepancy between the predicted output of the AI/ML model and the actual target. The goal of a loss function is to score and evaluate potential rules that a machine can learn from representation space. It defines an objective that can be minimized with respect to data mistakes made in a collection of data.

We can also score a loss function indirectly to evaluate a model's performance. This indirect approach is often beneficial for many reasons: optimization focus, non-intuitive scale, imbalanced data, and complex metrics. In order to address these issues, we can adopt validation metrics, hyperparameter tuning, model selection (comparing different models), and interpretability. These techniques can satisfy TAI properties in practice.

The bottom line is that writing score is much simpler than writing rules explicitly. However, if a mathematical model of a loss function on data becomes too complex, it can contribute to transparency issues. Many transparency issues typically arise from data-related challenges, such as data bias, data imbalance, labelling errors, noisy data, missing data, feature selection, data privacy and security, data distribution shift, and dataset size. Furthermore, when we intend to work out the loss function on data to satisfy TAI properties, many challenges lie in selecting the right algorithm to optimise loss functions.

## 2.3 Optimizer

Domingos [11] proposed five schools of thought on machine learning, and each school of thought mainly corresponds to one type of central problem. Table 1 illustrates the details of Domingos' five Schools of ML.

**Table 1.** Domingos Five Schools of Machine Learning

| Central Problem | Key Algorithms |
| --- | --- |
| Reasoning with symbols | Decision tree (if-then) |
| Analyzing perceptual information | Neural network/Deep neural network (perception) |
| Managing uncertainty | Bayesian networks (statistical data) |
| Discovering new structure | Genetic program (natural selection) |
| Exploiting similarities | Nearest Neighbours (previous cases) |

These five schools of ML provide a unifying approach for a broader understanding of the practical implications of algorithms, especially the selection of TAI properties in terms of robustness, reliability, reproducibility, and accuracy. Domingos touched on the strengths and weaknesses of each school's thought of ML.

### 2.3.1. Griadent Boost Machine

Historically, the decision tree method can be traced back to the Classification and Regression Tree (CART)[13] in the 1980s. The fundamental idea of a decision tree involves making inquiries for the given dataset and anticipating a precise prediction result. In contrast to other nonparametric algorithms, the decision tree method offers notable transparency and explanatory power for the prediction model [14]. Since then, it has evolved to bagging or bootstrap aggregating, random forest, and boosting iterations[15][16], including at least ten different boosting iteration techniques.

We can roughly divide the evolution history into four development phases: 1.) CART. 2.) Bagging bootstrap aggregation. 3.) Random Forests. 4.) Boosting iterations (See Fig. 3), although no clear demarcation line exists. We can consider the latter three phases as ensemble learning. The essence of ensemble learning is the "wisdom of crowds"[17] or meta-learning. Researchers have developed many boosting techniques. We can classify them into three classes: 1.) Adaptive boosting, which is the earliest algorithm. It is very slow in comparison to the next generation of models. 2.) Gradient Boosting Machine (GBM) is based on Frieman's idea of greedy function approximation [18], and 3.) Boosting models for particular types of datasets. The extreme Gradient Boost Machine (Xgbm) is the extension of GBM. The most compelling advantage of Xgbm is that we can run the algorithm in parallel on a high-performance computing (HPC) cluster or a cloud. Mathematically, we can use equations 2 and 3 to represent the gradient tree boosting algorithm for the predicted model:

$$f^* = \underset{f}{\mathrm{argmin}}\, L(f); \quad where\ f = \{f(x_i)\}_{i=1}^N\ ; L(f) = \sum_{i=1}^N L\,|y_i, f(x_i)| \quad (2)$$

$$f_B = \sum_{b=0}^B f_b,\ f_b \in \mathbb{R}^N; where\ f_b = f_{b-1} - \gamma g_b;\ g_b = \left\{\left[\frac{\partial L(f)}{\partial f}\right]_{f=f_{b-1}(x_i)}\right\}_{i=1}^N \quad (3)$$

Where $f^*$ is an optimal prediction function based on a genetic function $f$. $L(f)$ implies a loss function. $x_i\ (i = 1, 2, ... N)$ means "$i$" observation and $y_i$ stands for a predicted result. $f_B$ means the sum of "$B$" or the overall boosting functions based on N-features and $f_b$ represents a weak learner of boosting. $g_b$ is the steepest descent.

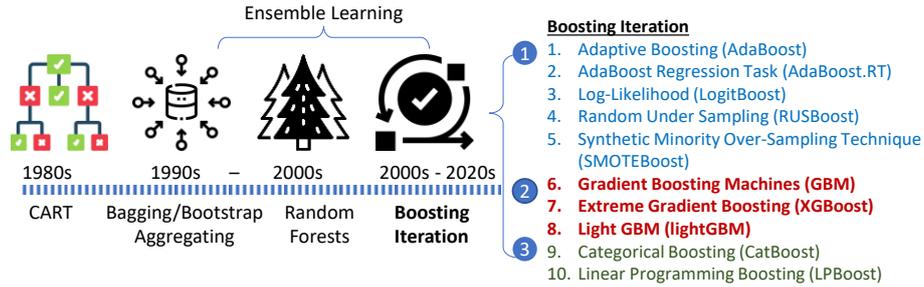

Fig. 3. Tree-Based Algorithms' Evolution

### 2.3.2. Transformer Models

Transformer [41], an attention-based structure, has generated significant interest due to its remarkable performance in computer vision (CV) [42] and natural language processing (NLP), exemplified by models like Generative Pre-trained Transformer (GPT) [43][44][45]. Its ability to model long-range dependencies and interactions in sequential data makes it an attractive option for time series modelling. Transformer models have been successfully applied in various time series forecasting tasks. State-of-the-art models include TimesNet [46], which extends 1D time series into 2D space and extracts complex temporal variations from transformed 2D tensors. Crossformer [47] embeds input data into a 2D vector array, utilizing cross-dimension dependency for multivariate time-series forecasting. PatchTST [48] introduces patching and channel-independent structures in their model, allowing for capturing local semantic information and benefiting from longer look-back windows. However, most of these models primarily focus on developing novel techniques to reduce the complexity of the original attention mechanism and achieve better performance. As a result, they are usually applied to energy, transport, and weather prediction applications. This research aims to evaluate these Transformer models from a Trustworthy AI perspective on the CDS dataset for strategic investment decisions.

### 2.4 CDS and Strategic Decision-Making

Merton [38] developed a distance-to-default (DTD) measure based on market information, assuming that the fundamental value of a firm follows a certain stochastic process and computing the default probability from the level and volatility of its market value. Das et al. [39] and Duan et al. [40] treat the default of a firm as an intensity process, $\lambda_t$; thus, the probability of surviving from starting time t=0 to default time t=τ is $s_\tau = \exp(-\int_0^\tau \lambda_t dt)$. The forward intensity $\lambda_t$ depends on the firm and economic features and is of exponential affine form,

$$\lambda_t = \exp[B'_{t-i} X_{t-i}], \ i \geq 0 \tag{4}$$

where $B_{t-i} = [\beta_{0(t-i)}, \ldots, \beta_{k(t-i)}]'$ is a vector of coefficients and $X_{t-i} = [1, X_{1(t-i)}, \ldots, X_{k(t-i)}]$ is a vector of features, including accounting-based, market-based, and macroeconomic variables ((such as equity value, price sale, inventory turnover, etc.). Assuming that condition on the given feature variables vector $X_{t-i}$, the forward default intensity is a constant, expressed as $E(\lambda_t | X_{t-i}) = \lambda$.

CDS enable market participants to shift the firm's default risk from an insurance buyer to an insurance seller. The buyer pays a premium to guarantee future potential protection. Hence, the decision of whether to buy or sell is often strategic because all market participants share the default risk. In order to predict selling or buying opportunities, the market participants require some trustworthy threshold level as an indicator. There are many accounting and economic features in a dataset; the challenge is to decide which feature is more important than the other and how to draw a threshold level. AI/ML can provide support for market participants' decisions.

### 2.5 Trustworthy AI (TAI) and Explainable AI (XAI)

The decision on TAI is very challenging, especially for an application of high-stakes decision, because it involves many aspects of subjective views, such as human beliefs, faith, experiences, ethical values, emotions, justice, fairness, equality, duty, right and wrong, and good and evil.[19] Many metrics are hard to quantify.

During the last decade, numerous ways of XAI have been developed because we often interpret AI/ML from different perspectives, such as users [20], logic [21] [22], biases [23], algorithms [24], responsibilities [25], methods/processes, models [26] [27], systems [28], stage [29], costs, and reasons [30]. Some researchers suggested that we should explain from a social science perspective [31]. Others [32] argue that it is not necessary to explain but interpret it. Burns et al. [33] proposed interpreting AI through hypothesis testing. However, Gilpin et al. [34] argued that the interpretation is insufficient. Whether we should explain or interpret it, many techniques are closely related. The issue of how we can apply a particular technique for a particular problem depends on a particular decision context and a dataset.

## 3  A Brid's Eye View of the Dataset and Model Environment

The dataset of the credit default swaps (CDS) has ten industrial sectors (See Fig. 4). The y-axis is "spread5" in the log scale. The spread5 represents the five-year contract of CDS. However, we increase spread5 by 10,000 times for analysis. It is a common practice for the CDS data.[52] The derivatives market usually has ten different CDS contracts regarding time or year [37]. The dataset that we have only has a five-year contract. The x-axis is the time domain between 3/Jan/2006 and 29/Dec/2017. It contains a total of 749,783 observations and 139 features. The data has been pre-cleaned manually. Consequently, many missing values have been deleted rather than estimated through multiple imputations.

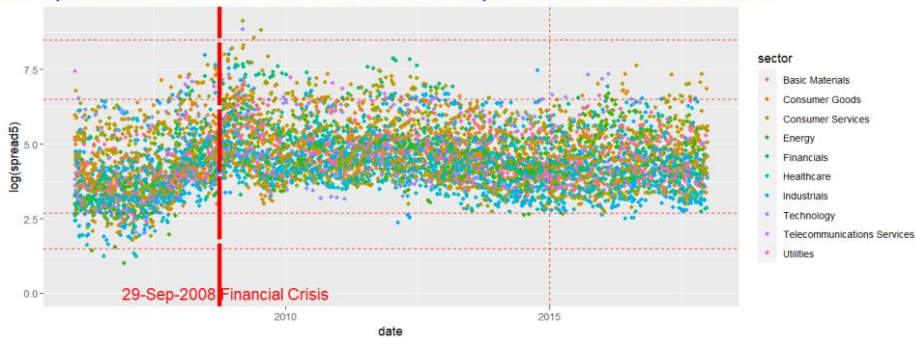

**Fig. 4.** Scatt Plot of 1% Sample of the Dataset

As shown in Fig.4, the overall fluctuation of spread5 price has been reduced from 1.5 - 8.5 before and during 2008 to 2.7 - 6.5 after 2015. However, we focus on the technology sector for this research.

### 3.1  Sub-dataset for Technology Sector

Compared with other industry sectors, the technology sector's fluctuation is relatively wider (See Fig.5) for 19 companies. However, the overall trend of spread5 contract price has been narrowed down after 2015. It is important to notice that the scatter plot (Fig.6) shows there are many missing values for some companies along the time domain. There could be many reasons why a company stopped trading for a while and resumed later.

The technology sector has 37,526 observations and 139 features. However, some features were generated during the pre-cleaning phase, which are dummy variables. Other variables have either no added values or are empty. Therefore, we removed these features and left with only 117 trainable features.

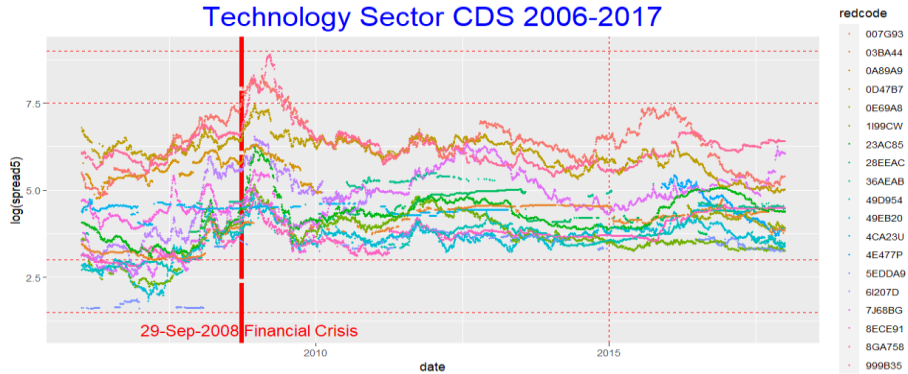

**Fig. 5.** Scatt Plot Sub-dataset for the Technology Sector

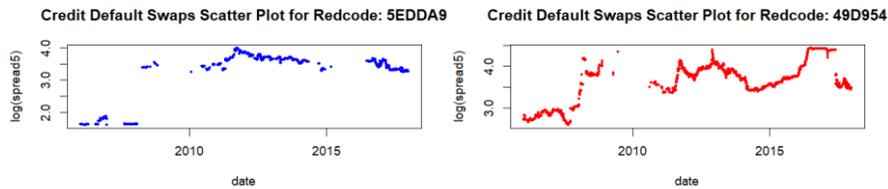

**Fig. 6.** Scatt Plot for Company's Redcode: 5EDDA9 and 49D954

### 3.2    Model Environment and Context

The experiment aims to develop a prediction model for strategic investment decisions of CDS contracts. We want AI/ML to generate the overall predictive model to support our investment decisions (buy or sell) by drawing a threshold level of some important metrics (features).

### 3.3    Selection of Algorithm for optimization

Based on the above scatter plots and the decision context, we select two types of predictive models, namely the decision tree-based and transformer models, for our experiments. The characteristics of tree-based models satisfy many TAI properties: better transparency, explanation, reproducibility, and reasoning. Fig.3 illustrates at least ten different tree-based models. During the initial phase of this study, we tried different tree-based models. The results indicate that Xgbm is the preferred model for prediction because Xgbm can run in parallel.

Compared with GBM, Xgbm shows its advantage in running hyperparameters searching for a large dataset if we run the algorithm on a high-performance computing (HPC) platform or a cloud. Xgbm is generally 8-10 times faster than the GBM. In addition, we do not have to worry about missing values and can aggregate results for all 19 companies.

## 4 Experimental Setup, Assumptions and Results

We first split the technology sub-dataset into a 70:30 ratio of 70% for training and 30% for testing. We also adopt a 5-fold cross-validation. The metric of the loss function is to measure root mean square error (RMSE). It is a common practice to use RMSE.[53]

We first ran GBM experiments and set up 36 grid points for the initial hyperparameter search to get a basic intuition about the terrain of the hyperparameter search field. Once this initial search has been done, we run a full-scale Xgbm hyperparameter search for 243 grid points on our HPC environment configured with a 128-core and 256 GB RAM cluster. And then, we will select the optimal parameters for the final prediction model.

After the final prediction model, we adopt five tools to explain the predictive model from global and local perspectives. These techniques include variable importance (VI) and partial dependent plots (PDP), individual conditional expectation (ICE), local interpretable model-agnostic explanations (LIME), and Shapley additive explanations (SHAP) values estimation. During our initial trial, we found that some categorical variables have a strong influence in the VI plot but very little explanatory power, such as "redcode" and "cusip". Therefore, we exclude these variables from our tests.

### 4.1 GBM Experimental Results

The first experiment is the GBM, which aims to have a rough estimation of some parameters, including the number of trees, shrinkage, interaction node depth, k value of cross-validation folds, bag fraction rate, and the number of minimum nodes.

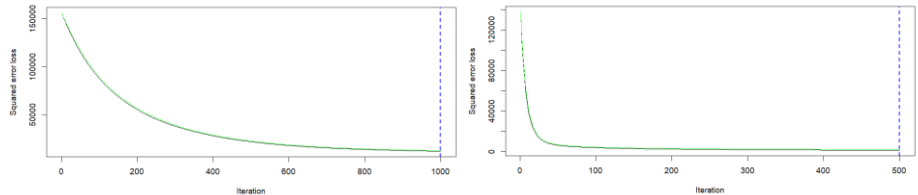

**Fig. 7.** GBM experimental results

The initial test showed that reducing the shrinkage rate (gradient step) does not help, but increasing interaction depth and the number of minimum nodes increases the prediction performance (See Fig.7). The right bag fraction value also increases performance (See Table 2). However, these parameters are not optimal. We have to run a hyperparameter search to find the optimal values of all parameters.

**Table 2.** GBM Experiment Results

| Parameters | Fig.7 Left Diagram | Fig.7 Right Diagram | Final Results |
|---|---|---|---|
| Distribution | Gaussian | Gaussian | Gaussian |
| # trees | 1000 | 500 | 800 |

| Shrinkage or learning rate | 0.01 | 0.1 | 0.3 |
|---|---|---|---|
| Interaction depth | 1 | 3 | 5 |
| # min. nodes | 1 | 3 | 5 |
| cv. fold | 5 | 5 | 5 |
| # predictors | 117 | 117 | 117 |
| Non-zero influence | 32 | 111 | 117 |
| Bag fraction | 1 | 1 | 0.85 |
| Train fraction | 1 | 1 | 1 |
| CPU usage time | 36.00 | 50.30 | 111.31 |
| System time | 0.47 | 0.56 | 0.20 |
| Elapsed time | 99.82 | 147.31 | 132.05 |
| RMSE | 112.548 | 46.372 | 29.512 |

### 4.2 Xgbm Experimental Results on HPC

We set up 243 grid points for the Xgbm hyperparameter search based on the intuition gained from initial tests. With a 128-node HPC cluster, it only takes 1.3 hours. We could achieve an even better RMSE of 25.97 by running a large hyperparameter (768 grid points) and more trees (3,500). However, the model improves very little for test RMSE after around 500 trees. It only improves the training RMSE. Therefore, we select 500 trees as a cutoff point.

**Table 3.** Xgbm Experiment Results

| Parameters | CPU usage time | System time | Elapsed time |
|---|---|---|---|
| HPC platform | 593,717.98 | 69.55 | 4,713.02 |
| Shrinkage or learning rate | Max tree depth | Min. rows /each end node | k fold CV |
| 0.10 | 5 | 1 | 5 |
| Subsample for each tree | Column sample | Number of trees | Min RMSE |
| 0.80 | 1 | 500 | 26.30 |

### 4.3 Transformer Models

We segmented the sub-dataset into 19 smaller datasets for the transformer model based on the company's code: "redcode". We then categorize these subsets into two groups for comparison according to nature threshold observations. In the experiments, we split each subset into a 70:10:20 ratio for training, validation, and testing. Afterwards, we employ three transformer models –TimesNet, PatchTST, and Crossformer for experiments on the HPC platform with one GPU and seven cores. The entire training process shows that PatchTST is a more efficient model. All transformer models in this experiment are in a "long-term forecasting" setting. The results are shown in Fig.8 and Table 4.

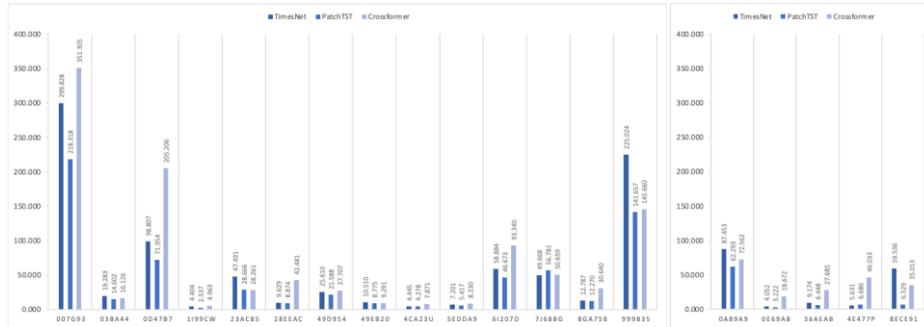

**Fig. 8.** 19 Companies of RMSE Results for CDS Prediction Models

The average RMSE for all models is 52.43. Based on the default parameters configuration, PatchTST's result is the best among these models. Notice that we did not implement a hyperparameter search for all transformer models because of the limited time and resources. All results are based on a random selection of the models' parameters. Therefore, the results are not optimal. Now, let us explain or interpret the prediction results.

**Table 4.** Transformer Models' RMSE Results

| Transformer Models | TimesNet | PatchTST | Crossformer | Average |
| --- | --- | --- | --- | --- |
| Training Time | 1203.64 | 368.19 | 1615.80 | 1062.54 |
| RMSE | 54.71 | 38.29 | 64.30 | 52.43 |

### 4.4 Variable Importance or Influence (VI) Results

The essence of the variable importance (VI) technique is its ability to identify and quantify the influence of individual features on the prediction performance. This technique is critical to understanding the most influential features in making accurate predictions. We plot the top 20 influence features or variables in Fig.9.

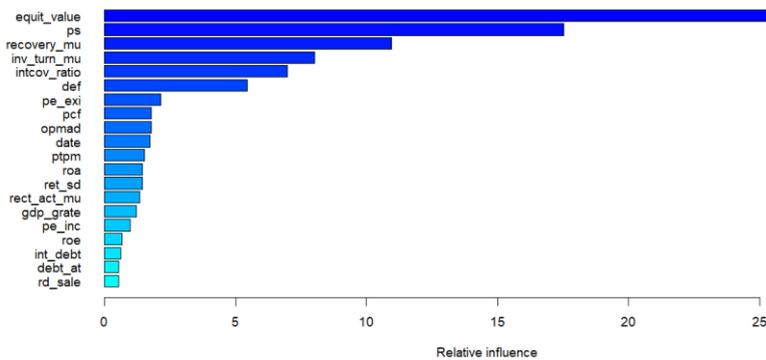

**Fig. 9.** VI results

Notice that the order of the top five influence features is relatively stable for all predictive models, but the rest of the features may change from one model to another. If the relative influence value is less than 10%, the ranking order of influence features will change.

### 4.5 Partial Dependent Plot (PDP) Results

According to Fig.9, which gives the variable importance (influence) results, we select the top six most relatively influential variables: "equity value"(total asset- total liabilities), "price sale" (market capitalization/total revenue), "recovery"(a kind of protection rate for a CDS buyer), "inventory turnover", "Interest coverage ratio", and "default spread" for PDP analysis (See Fig.10). The PDP provides a transparent and interpretable visualization of the relationship between a particular feature and the predictive outcome while keeping all other features constant. This technique assumes features are independent and identically distributed random variables (i.i.d.)

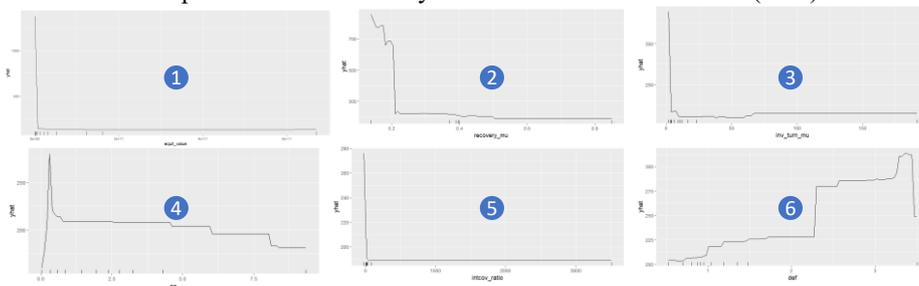

**Fig. 10.** PDP Results

For example, if the recovery value is less than the 0.2 threshold, the spread5 will drop nearly five times. (Refer to Fig.10. diagram 2). On the other hand, if the default spread value is larger than 2, the spread5 value increases by 0.5 base. This explainable technique exhibits the average view of prediction results. We can use the ICE technique to reveal the prediction results for more details of each instance.

### 4.6 Individual Conditional Expectation (ICE) Results

This study selects the top two variables (equity value and price sale) for the ICE experiments. There are two plots for each variable shown in Fig.11. One is a simple stack plot, and the other is a central plot. The ICE delivers a fine-grained understanding of how a specific feature affects the prediction of a single observation. To a certain extent, it provides a distributed view of a particular instance's influence on a particular feature prediction. This technique is invaluable for gaining insights into complex model behaviour and building trust in black-box model predictions. As indicated in Fig 3, the GBM is one type of ensemble model because more individual weak models are added to the ensemble. While ensemble models can be very powerful in predictive performance, they tend to be more complex than individual models. Balancing this complexity with the benefits of predictive accuracy is an important consideration when using ensembles in practice. Each black line is an observation, and the red line is a PDP.

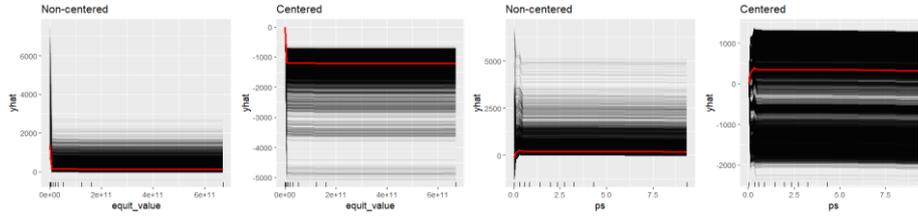

**Fig. 11.** ICE Results

### 4.7 Local interpretable model-agnostic explanations (LIME) Results

LIME is another technique that we tested in this paper. It aims to offer interpretable explanations for the predictions made by complex models. The primary goal of LIME is to make the predictive model more interpretable. It focuses on the local level rather than globally. Therefore, we selected eight individual cases for LIME analysis. (See Fig. 12). The first four cases (upper level) are before the 2008 financial crisis, and the other four cases (lower level) are after the 2008 financial crisis.

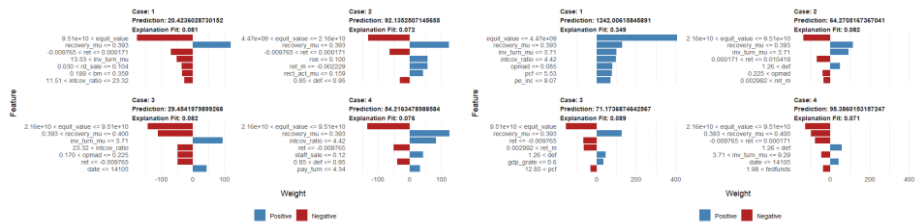

**Fig. 12.** LIME

Fig. 12 shows that the features contribute to the accuracy of a prediction. Only one case shows that "equity value" has a positive impact on the prediction value, but the predive RMSE is very large compared with other results. The remaining cases show a negative impact if RMSE values are less than 100.

### 4.8 Shapley Values (SHAP) Results

The Shapley values estimation attempts to explain complex machine learning models. It interprets some individual predictions. The essence of Shapley values implies the cooperative game theory and its application in allocating the values or contributions of each feature in a coalition game. Shapley values estimation captures fairness and marginality. It also considers the permutation of feature orderings and calculates each permutation's marginal contribution, then averages these contributions to estimate the Shapley value for each feature. (See Fig. 13)

However, data availability is one of the critical factors for Shapley values estimation. We might have explicit measurements of the feature's contributions. It depends on our decision context. Shapley value estimation is quite sensitive to distribution models. We adopt an empirical distribution model for the estimation (See Fig. 13). The result will be slightly different if we use a "copula" distribution. Shapley value estimation is both an art and a science. The explainable method depends on both data and the underlying characteristics of feature interaction. Overall, Shapley value estimation provides an equitable way to allocate each feature's contribution to the predicted case.

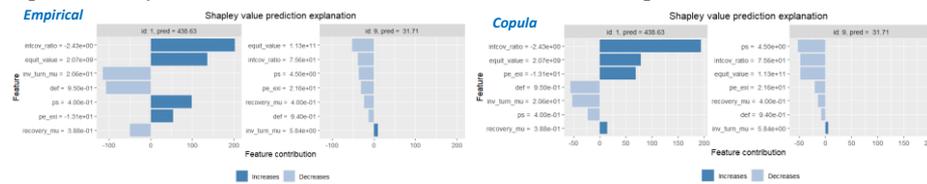

**Fig. 13.** SHAP Values

## 5      Results Analysis and Discussion

This paper's research question is "What kind of model (M), dataset (D), and algorithms (A) should satisfy the listed TAI properties (P) for strategic decision-making (E)?" In other words, how can we make the AI result to be trustworthy, which is how to decide what to decide? We proposed a framework consisting of three constituent components: representation space, loss function on data, and optimizer to satisfy all trustworthy properties. In this particular application, the issue is how to select a prediction model (representation space) of spread5 (or credit default swaps for a five-year contract) for a defined loss function (RMSE) on the pre-cleaned dataset, which includes 37,526 observations and 117 variables.

With the given dataset, we decided to use tree-based models and transformers for the experiments. Our experimental results illustrate that the Xgbm technique is more compelling than other models because it is very flexible for different datasets. We can also run the algorithm in parallel, even on a single machine with multiple cores. When we run a hyperparameter search (optimizer), Xgbm can save a lot of time. Table 3 illustrates that Xgbm can save as many as one week for 243 grid points hyperparameter searches. The Xgbm technique implies that we can find an optimal solution quickly that we can trust for a strategic investment decision.

Fig 8 demonstrates each company's RMSE result for different transformers: TimesNet, PatchTST, and Crossformer. PatchTST performs the best with a set of education guess parameters. However, compared with the XGBM model, the Xgbm model has more explanatory power. To better explain the CDS predictive model, we implemented five experiments to satisfy the listed trustworthy properties: transparency, explainable/ interpretability, usability, accuracy, robustness, reliability, and reproducibility. VI experiment demonstrates which features have a high influence on the predictive model. Based on the VI ranking order, we plot out the number of PDP that provides crucial insight for the strategic investment decision, which is when to sell or buy the CDS (spread5) contracts. The ICE plot shows how individual observation contributes to the overall PDP.

LIME provides a local explanation for the prediction model. It generates explanations by training an interpretable surrogate model (usually a simpler linear model) on a neighbourhood of the data point of interest. It tries to mimic the behaviour of the complex model locally. Generally, the LIME aims to make GBM or Xgbm more transparent and interpretable by generating local explanations of how a model arrived at a particular prediction for a specific instance. It is essential in applications where model interpretability is critical for trust and decision-making. Similarly, Shapley value estimation aims to quantify the contribution of each feature across all possible combinations of features. Shapley's method is often considered more stable and theoretically grounded, providing consistent explanations across different settings.

Compared with many previous research works [20][22][23][24][35], this study focuses on a systematic method to approach the TAI issue in the context of strategic investment decisions. We intend to provide a general framework for the TAI solution.

The limitation of this study is that we have not covered all TAI properties, such as data governance, privacy, and security issues. As we indicated before, the data has been pre-cleaned. While people cleaned the data, they deleted many observations due to missing values. This process may cause some issues with the accuracy of a prediction model. Furthermore, we did not run a hyperparameter search for transformer models. These issues will be a part of our future study when we receive the raw dataset and have enough computational resources. Another fundamental issue is that many AI/ML techniques focus only on correlation rather than logical reasoning.

Lenat and Marcus [49] argued that the Large Language Model (LLM) models are incomplete because they lack reasoning capabilities. Therefore, these models cannot be completely trustworthy. They proposed a rule-based system known as "Cyc" to be a complementary system for modern AI/ML models. They have been working on the "Cyc" project since 1984. Lenat and Marcus suggested that the modern AI/ML models are more like Kahneman's system-1 thinking [50], and "Cyc" is similar to Kahneman's system-2 thinking, which is underpinned by many logical reasoning approaches, such as inductive, reductive, and abductive methods. The Cyc project intends to build a common knowledge AI we can trust for strategic decision-making.

Steve Jobs once stated, "You cannot connect the dots looking forward; you can only connect them looking backwards."[51] Similarly, the modern AI/ML models can only extract the patterns of connected dots by looking backwards from a dataset, but strategic decision-making requires us to place dots by looking forward. It seems to be a dilemma or paradoxical. How can we trust the connected dots by looking backwards and lead to placing dots by looking forward? The answer could lie in the common knowledge of AI.

## 6 Conclusions and Future Direction

The research aims to create a novel framework for trustworthy AI from a strategic decision-making perspective. Based on the given dataset, we use GBM, Xgbm, and transformer models to test our hypothesis for the given dataset. The experimental results show that Xgbm is the compelling model for strategic investment decisions. This new framework of trustworthy AI provides a practical solution that can be applied to many contexts. It draws the baseline of deciding what to decide. Our main contribution is to build a bridge between trustworthy AI properties and practical ML solutions for strategic decision-making. However, we only cover a limited part of TAI in this research. We will cover all the TAI properties and the common knowledge AI for other decision contexts in future research.

## Acknowledgement

This research was funded in whole or part by the Luxembourg National Research Fund (FNR), grant ID C21/IS/16221483/CBD and grant ID 15748747. For open access, the author has applied a Creative Commons Attribution 4.0 International (CC BY 4.0) license to any Author Accepted Manuscript version arising from this submission.